\documentclass{article}
\usepackage[latin9]{inputenc}
\usepackage{geometry}
\usepackage{float}
\usepackage{amsmath}
\usepackage{amssymb}
\usepackage{graphicx}

\makeatletter

\providecommand{\tabularnewline}{\\}
\floatstyle{ruled}
\newfloat{algorithm}{tbp}{loa}
\providecommand{\algorithmname}{Algorithm}
\floatname{algorithm}{\protect\algorithmname}

\usepackage{subfigure}
\usepackage{times}

\usepackage{algorithm}
\usepackage[noend]{algpseudocode}
\algnewcommand\algorithmicinput{\textbf{Input:}}
\algnewcommand \Input{\item[\algorithmicinput]}
\algnewcommand\algorithmicoutput{\textbf{Output:}}
\algnewcommand \Output {\item[\algorithmicoutput]}
\algnewcommand\And{\textbf{and }}
\algnewcommand\Or{\textbf{or }}

\providecommand{\tabularnewline}{\\}

\usepackage{amsfonts}% blackboard math symbols

\usepackage{amsthm}

\usepackage{algorithm}

\newtheorem{defn}{Definition}\newtheorem{prop}{Proposition}\newtheorem{lem}{Lemma}

\newcommand{\desc}{d}
\newcommand{\pred}{a}
\newcommand{\edg}{\mathcal{E}}
\newcommand{\leaves}{\mathcal{L}}
\newcommand{\sums}{\mathcal{N}}

\title{
{ \textsc {\LARGE Learning Arbitrary Sum-Product Network Leaves with Expectation-Maximization}\footnote{Support of the German Science Foundation, grant GRK 1653, is gratefully acknowledged.}}
}

\author{
Mattia Desana${}^1$\thanks{Corresponding author. Email: mattia.desana@iwr.uni-heidelberg.de. }
$\,$ and Christoph Schnörr${}^{1,2}$.
}

\date{${}^1$Heidelberg Collaboratory for Image Processing (HCI)\\
${}^2$Image and Pattern Analysis Group (IPA)\\
Heidelberg University, Germany}

\makeatother

\begin{document}


\maketitle
\begin{abstract}
Sum-Product Networks with complex probability distribution at the
leaves have been shown to be powerful tractable-inference probabilistic
models. However, while learning the internal parameters has been amply
studied, learning complex leaf distribution is an open problem with
only few results available in special cases. In this paper we derive
an efficient method to learn a very large class of leaf distributions
with Expectation-Maximization. The EM updates have the form of simple
weighted maximum likelihood problems, allowing to use any distribution
that can be learned with maximum likelihood, even approximately. The
algorithm has cost linear in the model size and converges even if
only partial optimizations are performed. We demonstrate this approach
with experiments on twenty real-life datasets for density estimation,
using tree graphical models as leaves. Our model outperforms state-of-the-art
methods for parameter learning despite using SPNs with much fewer parameters. 
\end{abstract}

\section{Introduction}

Sum-Product Networks (SPNs, \cite{SPN2011}) are recently introduced
probabilistic models that possess two crucial characteristics: firstly,
inference in a SPN is always tractable; secondly, SPN enable to model
tractably a larger class of distributions than for Graphical Models
because they can model efficiently context specific dependences and
determinism (\cite{Boutilier96context-specificindependence}). Due
to their ability to use exact inference in complex distributions SPNs
are state-of-the-art models in density estimation (see e.g. \cite{SPNstructureLearning2013,Rahman16mergeSPN})
and have been successfully used in computer vision (\cite{SPNlangMod2014},
\cite{SPNspeechPeharz2014}, \cite{SPnforActRecogn2015}).

\begin{figure}[t]
\center \includegraphics[width=0.4\textwidth]{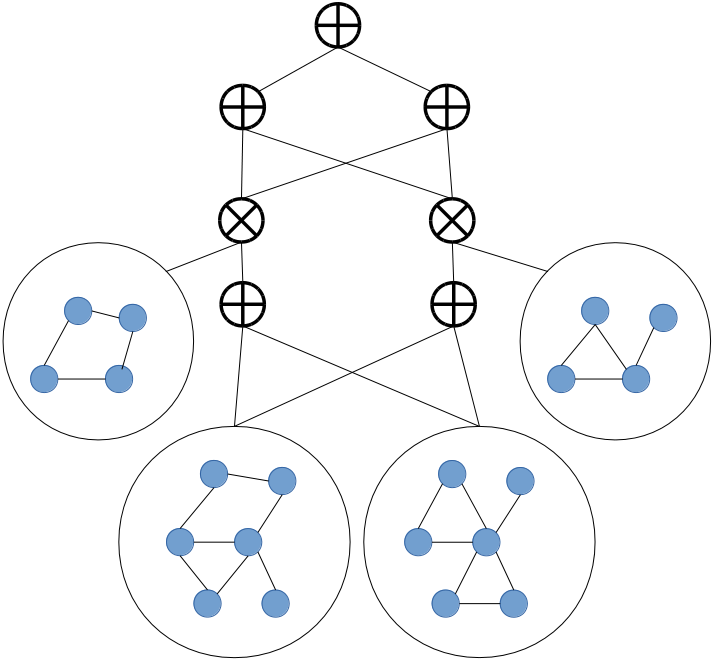} \caption{An example SPN with complex leaf distributions (in this case, probabilistic
graphical models). \label{fig:Top:-SPN-representing}\label{fig:SPNtrees}}
\end{figure}

SPNs are modelled by a directed acyclic graph with two sets of parameters:
edge coefficients at internal sum nodes and probabilistic distributions
at the leaves. Most SPN models use very simple leaf models in form
of indicator variables. However, using leaf distributions with complex
structure allows to create SPNs with high modelling power and flexibility,
as shown for instance in \cite{Rahman16} using tree graphical models
as leaves and in \cite{SPnforActRecogn2015} using Bag-of-Words (example:
fig. \ref{fig:SPNtrees}). While there are several methods to learn
edge coefficients (\cite{SPNdiscrimLearning2012,zhao16,Zhao16collapsVarInf}),
learning the leaf distribution parameters is still an open problem
in the general case. The only method we are aware of is \cite{peharz15},
which works for the special case of univariate distribution in the
exponential family (although the authors suggest it can be extended
to the multivariate case).

The goal of this paper is to learn leaf distributions with complex
structure in a principled way. To do so, we obtain a novel derivation
of Expectation-Maximization for SPNs that allows to cover leaf distribution
updates (section \ref{sec:Expectation-Maximization}). The first step
in this derivation is providing a new theoretical result relating
the SPN and a subset of its encoded mixture (Proposition \ref{prop:Consider-a-SPN}
in the following). Exploiting this new result EM for SPN leaves assumes
the form of a \emph{weighted maximum likelihood} problem, which is
a slight modification of standard maximum likelihood and is well studied
for a wide class of distributions. The algorithm has computational
cost \emph{linear} in the number of SPN edges.

\emph{Convergence} of the algorithm is guaranteed as long as the maximization
is even \emph{partially} performed. Therefore, any distribution where
at least an approximate log-likelihood maximization method is available
can be used as SPN leaf. This result allows to use a \emph{very wide}
family of leaf distributions and train them efficiently and straightforwardly.
Particularly nice results hold when leaves belong to the exponential
family, where the M-step has a single optimum and the maximization
can often be performed efficiently in closed form.

To test the potential advantages of training complex leaves we perform
experiments on a set of twenty widely used datasets for density estimation,
using a SPN with tree graphical models as leaf distributions (section
\ref{sec:Application---Soft}). We show that a simple SPN with tree
graphical model leaves learned with EM state-of-the-art methods for
parameter learning while using much smaller models. These results
suggests that much of the complexity of the SPN structure can be encoded
in complex, trainable leaves rather than in a large number of edges,
which is a promising direction for future research.

%\begin{figure}[t]
%\center\includegraphics[width=0.25\textwidth]{img/SPN_trees}\caption{A SPN with tree graphical models as leaves. The SPN \emph{weights}
%and the \emph{structure} and \emph{potentials} of the trees can be
%learnt \emph{jointly} and efficiently with our derivation of EM. \label{fig:SPNtrees}}
%\end{figure}

%\begin{figure}[t]
%\center
%\includegraphics[width=0.8\textwidth]{img/1plus2}
%\caption{
%(Left): Parity distribution from \cite{SPN2011}. Top: SPN representing the
%uniform distribution over states of five variables containing an even
%number of $1$\textquoteright{}s. Bottom: mixture model for the same
%distribution. (Right): A SPN with tree graphical models as leaves. The SPN \emph{weights}
%and the \emph{structure} and \emph{potentials} of the trees can be
%learnt \emph{jointly} and efficiently with our derivation of EM.
%\label{fig:Top:-SPN-representing}
%}
%\end{figure}

%%%%%%%%%%%%%%%%%%%%%%%%%%%%%%%%%%%%%%%%%%%%%%%%%%%%%%

\section{Sum-Product Networks \label{sec:Sum-Product-Networks}}

We start with the definition of SPN based on \cite{SPNstructureLearning2013}.
Let $X$ be a set of random variables, either continuous or discrete.
\begin{defn} \label{def:SPN} Sum-Product Network (SPN) : \end{defn} 
\begin{enumerate}
\item \label{enu:spn1}A tractable distribution $\varphi(X)$ is a SPN $S(X)$. 
\item \label{enu:spn2}The product $\prod_{k}S_{k}(X_{k})$ of SPNs $S_{k}(X_{k})$
is a SPN $S\left(\bigcup_{k}X_{k}\right)$ if all sets $X_{k}$ are
disjoint. 
\item \label{enu:spn3}The weighted sum $\sum_{k}w_{k}S_{k}(X)$ of SPNs
$S_{k}(X)$ is a SPN $S(X)$ if the weights $w_{k}\in\mathbb{R}$
are nonnegative (notice that $X$ is in common for each SPN $S_{k}$). 
\end{enumerate}
By associating a node to each product, sum and tractable distribution
and adding edges between an operator and its inputs, a SPN can be
represented as a rooted Directed Acyclic Graph (DAG) $\mathcal{G}(\mathcal{V},\mathcal{E})$
with sums and products as internal nodes and tractable distributions
as leaves (example: fig. \ref{fig:spnPath}). This definition generalizes
SPNs with indicator variables as leaves, since indicator variables
are a special case of discrete distribution where the probability
mass completely lies on a single variable state. A SPN is \emph{normalized
}if weights of outgoing edges of sum nodes sum to $1$: $\sum_{k}w_{k}=1$.
We consider \emph{only normalized SPNs}, without loss of generality
(\cite{Peharz2015phd}).

\textbf{Notation. } We use the following notation throughout the paper.
Let $X$ be a set of variables (either continuous or discrete depending
on the context) and let $x$ be an assignment of these variables.
$S_{q}\left(X_{q}\right)$ denotes the sub-SPN rooted at a node $q$
of $S$, with $X_{q}\subseteq X$. $S\left(x\right)$ denotes the
evaluation of $S$ with assignment $x$ (see below), and $\frac{\partial S\left(x\right)}{\partial S_{q}}$
is the derivative of $S(x)$ w.r.t. node $q$. The term $\varphi_{l}\left(X_{l}\right)$
denotes the distribution of leaf node $l$. $ch(q)$ and $pa(q)$
denote the children and parents of $q$ respectively, and $(q,i)$
indicates an edge between $q$ and its child $i$, associated to a
weight $w_{i}^{q}$ if $q$ is a sum node. Finally, let $\mathcal{E}(S)$,
$\mathcal{V}(S)$ and $\mathcal{\leaves}\left(S\right)$ denote respectively
the set of edges, nodes and leaves in $S$.

\textbf{Parameters. } Let $W$ denote the set of sum node weights
and let $\theta$ denote the set of parameters governing the leaf
distributions. We write $S\left(X|W,\theta\right)$ to explicitly
express dependency of $S$ on these parameters. Each leaf distribution
$\varphi_{l}$ is associated to a parameter set $\theta_{l}\subseteq\theta$.
For instance, $\theta_{l}$ contains mean and covariance for Gaussian
leaves, and tree structure and potentials for tree graphical model
leaves.

\textbf{Evaluation.}\label{subsec:Evaluation} The evaluation of $S(X)$
for evidence $x$, written $S(x)$, proceeds by first evaluating the
leaf distributions with assignment $x$, then evaluating each internal
node from the leaves to root and taking the value of the root. Evaluating
any valid SPN corresponds to evaluating a probability distribution
(\cite{SPN2011}). Computing $S(x)$, the partition function and the
quantities $S_{q}\left(x\right)$ and $\frac{\partial S\left(x\right)}{\partial S_{q}}$
for \emph{each} node in $S$ requires performing a single up-and-down
pass over all network nodes and has a $O\left(|\edg|\right)$ time
and memory cost (\cite{SPN2011}).

%%%%%%%%%%%%%%%%%%%%%%%%%%%%%%%%%%%%%%%%%%

\section{SPNs as Mixture Models \label{sec:SPNs-as-aMix}}

\begin{figure}
\center\includegraphics[width=0.4\textwidth]{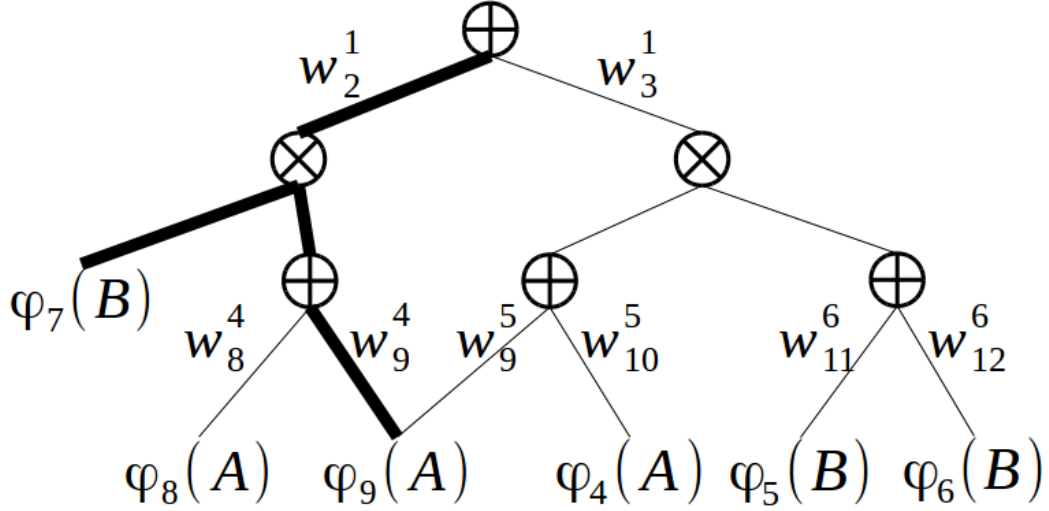}\caption{A SPN $S\left(A,B\right)$ in which a subnetwork $\sigma_{c}$ of
$S$ is highlighted. This subnetwork corresponds to a coefficient
$\lambda_{c}=w_{2}^{1}w_{9}^{4}$ and component $P_{c}\left(A,B\right)=\varphi_{7}\left(B\right)\varphi_{9}\left(A\right)$.\label{fig:spnPath}}
\end{figure}

This section discusses the interpretation of SPNs as a mixture model
derived in \cite{DennisVentura15} and \cite{zhao16}, on which we
will base our derivation of EM. \begin{defn} A \emph{subnetwork}
$\sigma_{c}$ of $S$ is a SPN constructed by first including the
root of $S$ in $\sigma_{c}$, then processing each node $q$ included
in $\sigma_{c}$ as follows:\end{defn} 
\begin{enumerate}
\item \vspace{-5pt}
 If $q$ is a sum node, include in $\sigma_{c}$ one child $i\in ch\left(q\right)$
with relative weight $w_{i}^{q}$. Process the included child. 
\item \vspace{-5pt}
 If $q$ is a product node, include in $\sigma_{c}$ all the children
$ch\left(q\right)$. Process the included children. 
\item \vspace{-5pt}
 If $q$ is a leaf node, do nothing. 
\end{enumerate}
Example: fig. \ref{fig:spnPath}. Any subnetwork \emph{is a} \emph{tree}
%(\cite{DennisVentura15})%
. Let $C={\sigma_{1},\sigma_{2},...,\sigma_{C}}$ be the \emph{number
of different subnetworks} obtainable from $S$ for different choices
of included sum node children. The number of subnetworks $C$ can
be \emph{exponentially larger} than the number of edges $|\edg|$.
\begin{defn} \label{def:pcLambda} For a subnetwork $\sigma_{c}$
of $S\left(X|W,\theta\right)$ we define a \emph{mixture coefficient
} $\lambda_{c}=\prod_{\left(q,j\right)\in\mathcal{\edg}\left(\sigma_{c}\right)}w_{j}^{q}$
and a \emph{mixture component} $P_{c}(X|\theta)=\prod_{l\in\leaves\left(\sigma_{c}\right)}\varphi_{l}\left(X_{l}|\theta_{l}\right)$.
\end{defn} Note that mixture coefficients are products of sum weights
and mixture components are products of leaves in $\sigma_{c}$ (fig.
\ref{fig:spnPath}). \begin{prop} \label{prop:A-SPN-S}$S|W,\theta\left(X\right)$
represents the following mixture model: 
\begin{equation}
S\left(X|W,\theta\right)=\sum_{c=1}^{C}\lambda_{c}\left(W\right)P_{c}\left(X|\theta\right)\label{eq:spnmix}
\end{equation}
\end{prop} Proof: see \cite{DennisVentura15}. Notice that since
$C\gg|\edg|$, it follows that a SPN encodes a mixture which can be
intractably large if explicitly represented.

\begin{figure}
\center\includegraphics[width=0.4\textwidth]{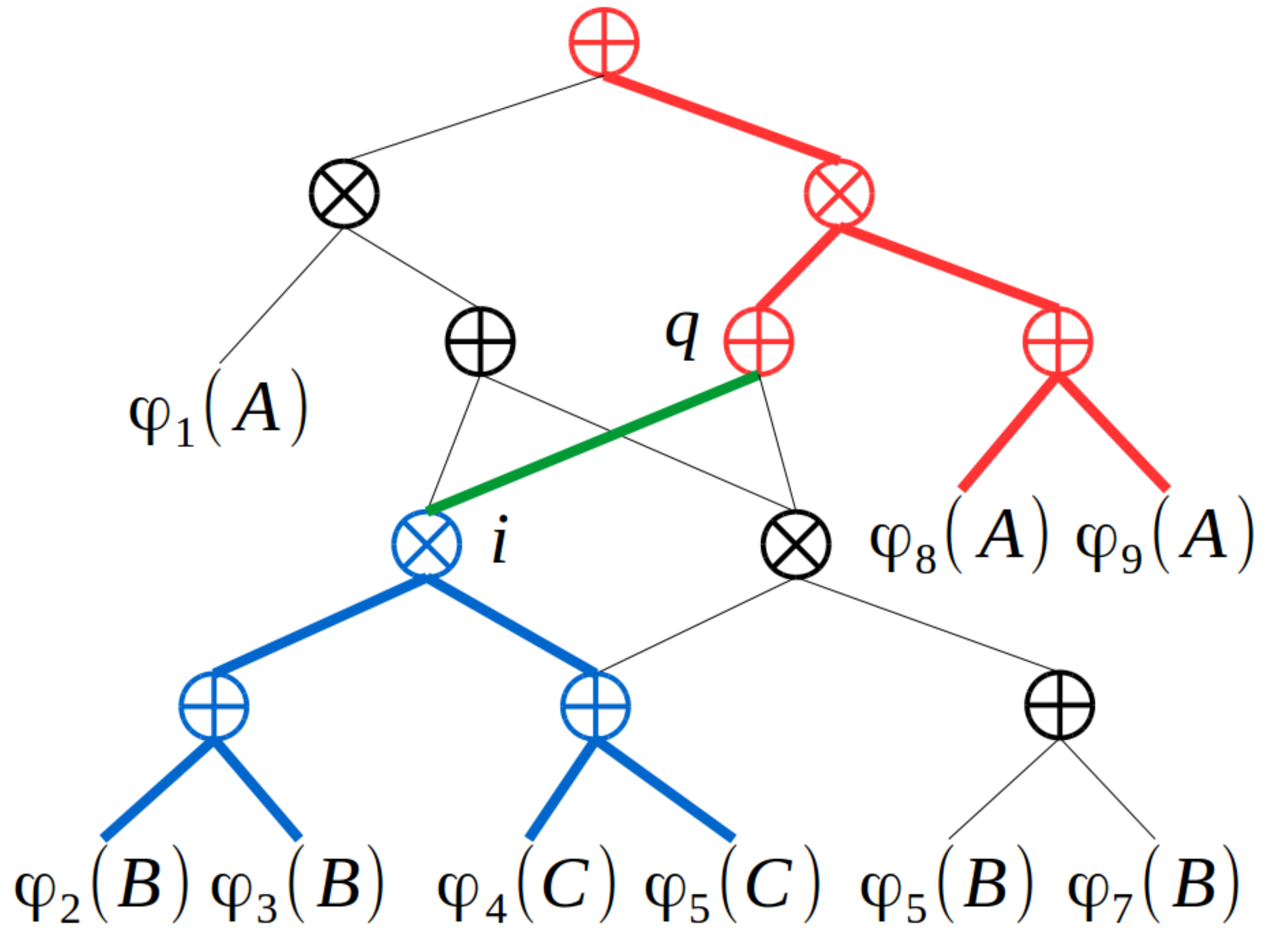}\caption{Visualization of Proposition \ref{prop:Consider-a-SPN}. The colored
part is the set of edges traversed by all subnetworks crossing $\left(q,i\right)$.
The blue part represents $S_{i}$ and the red part $\frac{\partial S\left(X\right)}{\partial S_{q}}$.\label{fig:lemEx2}}
\end{figure}

%\begin{figure}
%\center\includegraphics[width=0.48\textwidth]{img/lemmaExDouble}\caption{Visualization of Lemma \ref{prop:Consider-a-SPN}. Left: a subnetwork
%$\sigma_{c}$ crossing $\left(q,i\right)$. Right: The colored part
%is the set of edges traversed by all subnetworks crossing $\left(q,i\right)$.
%The blue part represents $S_{i}$ and the red part covers terms appearing
%in $\frac{\partial S\left(X\right)}{\partial S_{q}}$. \label{fig:lemEx2}}
%\end{figure}

We now introduce a \emph{new result} that is crucial for our derivation
of EM, reporting it here rather than in the proofs section since it
contributes to the set of analytical tools for SPNs. \begin{prop}
\label{prop:Consider-a-SPN}Consider a SPN $S(X)$, a sum node $q\in S$
and a node $i\in ch(q)$. The following relation holds: \vspace{-5pt}
 
\begin{equation}
\sum_{k:\left(q,i\right)\in\edg\left(\sigma_{k}\right)}\lambda_{k}P_{k}\left(X\right)=w_{i}^{q}\frac{\partial S\left(X\right)}{\partial S_{q}}S_{i}\left(X\right)\label{eq:sfsfr}
\end{equation}
where $\sum_{k:\left(q,i\right)\in\edg\left(\sigma_{k}\right)}$ denotes
the sum over all the subnetworks $\sigma_{k}$ of $S$ that include
the edge $\left(q,i\right)$. \end{prop} Proof: in Appendix \ref{subsec:Proof-of-Lemma}.
This result states that the value of each sub-mixture composed by
all the subnetworks crossing $\left(q,i\right)$, which has potentially
intractable large size, can be evaluated in constant time after having
evaluated and derivated $S(x)$ once. This results is crucial in the
derivation of EM (Appendix \ref{appendix}) where we need to evaluate
such subsets of solutions repeatedly. Note also that $\sum_{k:\left(q,i\right)\in\edg\left(\sigma_{k}\right)}\lambda_{k}P_{k}\left(X\right)$
corresponds to the evaluation of a non-normalized SPN which is a subset
of $S$ - e.g. the colored part in fig. \ref{fig:lemEx2}.

%%%%%%%%%%%%%%%%%%%%%%%%%%%%%%%%%%%%%%%%%%%%%%%%%%%%%%%%%%%%%%%%%%%%%%

\section{Expectation Maximization \label{sec:Expectation-Maximization}}

In this section we obtain a novel derivation of Expectation-Maximization
for SPNs by directly applying EM for mixture models to the exponentially
large mixture encoded by a SPNs exploiting Proposition \ref{prop:Consider-a-SPN}.
We obtain a procedure to learn SPNs with a broad class of leaf distributions,
and show that the algorithm converges under mild conditions.

Expectation Maximization is an elegant and widely used method for
finding maximum likelihood solutions for models with latent variables
(see e.g. \cite[11.4]{Murphy:2012}). Given a distribution $P(X)=\sum_{c=1}^{C}P\left(X,c|\pi\right)$
where $c$ are latent variables and $\pi$ are the distribution parameters
our objective is to maximize the log likelihood $\sum_{n=1}^{N}\ln\sum_{c=1}^{C}P(x_{n},c|\pi)$
over a dataset of observations $\left\{ x_{1},x_{2},...,x_{N}\right\} $.
EM proceeds by updating the parameters iteratively starting from some
initial configuration $\pi_{old}$. An update step consists in finding
$\pi^{*}=\arg\max_{\pi}Q\left(\pi|\pi_{old}\right)$, where 
\[
Q\left(\pi|\pi_{old}\right)=\sum_{n=1}^{N}\sum_{c=1}^{C}P\left(c|x_{n},\pi_{old}\right)\ln P\left(c,x_{n}|\pi\right).
\]
We want to apply EM to the mixture encoded by a SPN, which is in principle
intractably large. First, using the relation between SPN and encoded
mixture model in Proposition \ref{prop:A-SPN-S} we identify 
\begin{align*}
P(c,x_{n}|\pi) & =\lambda_{c}\left(W\right)P_{c}\left(x_{n}|\theta\right),\\
P(x_{n}|\pi_{old}) & =S\left(x_{n}|W_{old},\theta_{old}\right),
\end{align*}
therefore:
\[
P(c|x_{n},\pi_{old})=P(c,x_{n}|\pi_{old})/P(x_{n}|\pi_{old})=\lambda_{c}\left(W_{old}\right)P_{c}\left(x_{n}|\theta_{old}\right)/S\left(x_{n}|W_{old},\theta_{old}\right).
\]
Applying these substitutions and dropping the dependency on $W_{old},\theta_{old}$
for compactness, $Q\left(W,\theta|W_{old},\theta_{old}\right)$ becomes:
\begin{equation}
Q\left(W,\theta\right)=\sum_{n=1}^{N}\sum_{c=1}^{C}\frac{\lambda_{c}P_{c}\left(x_{n}\right)}{S\left(x_{n}\right)}\ln\lambda_{c}\left(W\right)P_{c}\left(x_{n}|\theta\right)\label{eq:Q}
\end{equation}
which we maximize for $W$ and $\theta$ in the following sections.
%\begin{align}
%Q\left(W,\theta\right) &= \sum_{n=1}^{N}\sum_{c=1}^{C}\frac{\lambda_{c}P_{c}\left(x_{n}\right)}{S\left(x_{n}\right)}\ln\lambda_{c}\left(W\right)P_{c}\left(x_{n}|\theta\right) \\
%&= 
%\underbrace{\sum_{n=1}^{N}\sum_{c=1}^{C}\frac{\lambda_{c}P_{c}\left(x_{n}\right)}{S\left(x_{n}\right)}\ln\lambda_{c}\left(W\right)}_{ Q_W \left(W\right) }
%+ 
%\underbrace{\sum_{n=1}^{N}\sum_{c=1}^{C}\frac{\lambda_{c}P_{c}\left(x_{n}\right)}{S\left(x_{n}\right)}\ln P_{c}\left(x_{n}|\theta\right)}_{ Q_{\theta}\left(\theta\right) }
%\label{eq:Q}
%\end{align}
%In the following sections we will maximize $Q\left(W,\theta\right) = Q_W \left(W\right)+Q_{\theta}\left(\theta\right)$ for $W$ and $\theta$. 

\subsection{Edge Weights Update}

We begin with EM updates for weights. Simplifying $Q\left(W,\theta\right)$
through the use of Proposition \ref{prop:Consider-a-SPN} (Appendix
\ref{subsec:EM-wDeriv}) the objective function for $W$ becomes:
\begin{align}
W^{*} & =\arg\max_{W}Q_{W}\left(W\right)\label{eq:wup}\\
Q_{W}\left(W\right) & =\sum_{q\in\sums\left(S\right)}\sum_{i\in ch\left(q\right)}\beta_{i}^{q}\ln w_{i}^{q}\\
\beta_{i}^{q} & =w_{i,old}^{q}\sum_{n=1}^{N}S\left(x_{n}\right)^{-1}\frac{\partial S\left(x_{n}\right)}{\partial S_{q}}S_{i}\left(x_{n}\right)\nonumber 
\end{align}
The evaluation of terms $\beta_{i}^{q}$, which depend only on $W_{old},\theta_{old}$
and are therefore constants in the optimization, is the \emph{E step}
of the EM algorithm. We now maximize $Q_{W}\left(W\right)$ subject
to $\sum_{i}w_{i}^{q}=1\forall q\in\sums\left(S\right)$ (\emph{M
step}).

\textbf{Non shared weights.} If weights at each node $q$ are disjoint,
then we can move the $\max$ inside the sum, obtaining separated maximizations
each in the form $\arg\max_{w^{q}}\sum_{i\in ch(q)}\beta_{i}^{q}\ln w_{i}^{q}$,
where $w^{q}$ is the set of weights outgoing from $q$. Now, the
same maximum is attained multiplying by $k=\frac{1}{\sum_{i}\beta_{i}^{q}}$.
Then, defining $\bar{\beta}_{i}^{q}=k\beta_{i}^{q}$ , we can equivalently
find $\arg\max_{w^{q}}\sum_{i\in ch(q)}\bar{\beta}_{i}^{q}\ln w_{i}^{q}$,
where $\bar{\beta}_{i}^{q}$ is positive and sums to $1$ and therefore
can be interpreted as a discrete distribution. This is then the maximum
of the cross entropy $\arg\max_{w^{q}}\left(-\mathbb{H}\left(\bar{\beta}_{i}^{q},w_{i}^{q}\right)\right)$
defined e.g. in \cite[2.8.2]{Murphy:2012}, attained for $w_{i}^{q}=\bar{\beta}_{i}^{q}$,
which corresponds to the following update: 
\begin{equation}
w_{j}^{q*}=\beta_{j}^{q}/\sum_{i}\beta_{i}^{q}.\label{eq:wup-1}
\end{equation}
This is the same weight update obtained with radically different approaches
in \cite{peharz15,zhao16}.

\textbf{Shared weights. } Our derivation of weights updates allows
a straightforward extension to the case of shared-weights nodes. Weights
shared between different sum nodes appear for instance in convolutional
SPNs (see e.g. \cite{SPNlangMod2014}). To keep notation simple let
us consider only two nodes $q_{1},q_{2}$ constrained to share weights,
that is $w_{i}^{q_{1}}=w_{i}^{q_{2}}=\hat{w}$ for every child $i$,
where $\hat{w}$ is the set of shared weights. We then rewrite $Q_{W}\left(W\right)$
insulating the part depending on $\hat{w}$ as $Q_{W}\left(W\right)=\sum_{i\in ch(q)}\beta_{i}^{q_{1}}\ln w_{i}^{q_{1}}+\sum_{i\in ch(q)}\beta_{i}^{q_{2}}\ln w_{i}^{q_{2}}+const$
(the constant includes terms not depending on $w^{q}$). Then, employing
the weight sharing constraint, maximization of $Q_{W}$ for $\hat{w}$
becomes $\arg\max_{\hat{w}}\sum_{i\in ch(q)}\left(\beta_{i}^{q_{1}}+\beta_{i}^{q_{2}}\right)\ln\hat{w}$.
As in the non-shared case, we end up maximizing the cross entropy
$-\mathbb{H}\left(k\left(\beta_{i}^{q_{1}}+\beta_{i}^{q_{2}}\right),\hat{w}_{i}\right)$.
Generalizing for an arbitrary set of nodes $D$ such that any node
$q\in D$ has shared weights $\hat{w}$, the weight update for $\hat{w}$
is as follows: 
\begin{equation}
\hat{w}_{j}^{*}=\frac{\sum_{q\in D}\beta_{j}^{q}}{\sum_{i}\sum_{q\in D}\beta_{i}^{q}}.\label{eq:wup-1-1}
\end{equation}

\subsection{Leaf Distribution Updates}

We now consider learning leaf distributions. Simplifying $Q\left(W,\theta\right)$
through the use of Proposition \ref{prop:Consider-a-SPN} (Appendix
\ref{subsec:EM-thetaDeriv}), the objective function for $\theta$
becomes: 
\begin{align}
\theta^{*} & =\arg\max_{\theta}Q_{\theta}\left(\theta\right)\label{eq:aasda22}\\
Q_{\theta}\left(\theta\right) & =\sum_{l\in\leaves\left(S\right)}\sum_{n=1}^{N}\alpha_{ln}\ln\varphi_{l}\left(x_{n}|\theta_{l}\right)\label{eq:aasda223}\\
\alpha_{ln} & =S\left(x_{n}\right)^{-1}\frac{\partial S\left(x_{n}\right)}{\partial S_{l}}S_{l}\left(x_{n}\right)\nonumber 
\end{align}
The evaluation of terms $\alpha_{ln}$, which are constant coefficients
in the optimization since they depend only on $W_{old},\theta_{old}$,
is the \emph{E step} and can be seen as computing the \emph{responsibility}
that leaf distribution $\varphi_{l}$ assigns to the $n$-th data
point, just as in EM for classical mixture models. Importantly, we
note that the maximization eq. \ref{eq:aasda22} is \emph{concave}
as long as $\ln\varphi_{l}\left(X_{l}|\theta\right)$ is concave,
in which case there is an \emph{unique global optimum}. Also note
that normalizing $\alpha_{l}$ in eq. \ref{eq:aasda223} dividing
each $\alpha_{ln}$ by $\sum_{n}\alpha_{ln}$ we attain the same maximum
and avoid numerical problems due to very small values.

\textbf{Non shared parameters. }Introducing the hypothesis that parameters
$\theta_{l}$ are disjoint at each leaf $l$, we obtain separate maximizations
in the form: \vspace{-5pt}
 
\begin{equation}
\theta_{l}^{*}=\arg\max_{\theta_{l}}\sum_{n=1}^{N}\alpha_{ln}\ln\varphi_{l}\left(x_{n}|\theta_{l}\right)\label{eq:aasda}
\end{equation}
In this formulation one can recognize a \emph{weighted maximum likelihood}
problem, where each data sample $n$ is weighted by a soft-count coefficient
$\alpha_{ln}$.

\textbf{Shared parameters.} Let us consider two leaf nodes $k,j$
associated to distributions $\varphi_{k}(X_{k}|\theta_{k}),\varphi_{j}(X_{j}|\theta_{j})$
respectively, such that $\theta_{k}=\theta_{j}=\hat{\theta}$ are
shared parameters. Eq. \ref{eq:aasda22} for $k,j$ becomes %\begin{align}
\[
Q_{\theta}\left(\theta\right)=\sum_{n=1}^{N}\alpha_{kn}\ln\varphi_{k}(x_{n}|\hat{\theta})+\sum_{n=1}^{N}\alpha_{jn}\ln\varphi_{j}(x_{n}|\hat{\theta})+const(\hat{\theta}).
\]
Generalizing to an arbitrary set of leaves $D$ such that each leaf
$l\in D$ has a distribution $\varphi\hat{X_{l}|\theta}$ and dropping
the constant term, we obtain: 
\begin{equation}
\hat{\theta}^{*}=\arg\max_{\hat{\theta}}\sum_{l\in D}\left(\sum_{n=1}^{N}\alpha_{ln}\ln\varphi_{l}(x_{n}|\hat{\theta})\right).\label{eq:aasdaShared}
\end{equation}
The objective function now contains a sum of logarithms, therefore
it cannot be maximized as separate problems over each leaf as in the
non shared case. However, it is still \emph{concave} in $\theta$
as long as $\ln\varphi_{l}\left(X_{l}|\theta\right)$ is concave,
in which case there is an \emph{unique global optimum} (this holds
for exponential families, discussed next). Then, the optimal solution
can be found with iterative methods such as gradient descent or second
order methods.

\paragraph{Exponential Family Leaves.\label{subsec:Exponential-Family-Leaves}}

For distributions in the exponential family eq. \ref{eq:aasda22}
is \emph{concave} and therefore a \emph{global optimum} can be reached
(see e.g. \cite[11.3.2]{Murphy:2012}). Additionally, the solution
is often available efficiently in \emph{closed form}. Let us consider
two relevant examples. If $\varphi_{l}\left(X_{l}\right)$ is a \emph{multivariate
Gaussian} $\mathcal{N}(\mu_{l},\Sigma_{l})$, the solution of eq.
\ref{eq:aasda} is obtained e.g. in \cite[11.4.2]{Murphy:2012} as
$r_{l}=\sum_{n=1}^{N}\alpha_{ln}$, $\mu_{l}=\frac{\sum_{n=1}^{N}\alpha_{ln}x_{n}}{r_{l}}$
and $\Sigma_{l}=\frac{\sum_{n=1}^{N}\alpha_{ln}\left(x_{n}-\mu_{l}\right)\left(x_{n}-\mu_{l}\right)^{T}}{r_{l}}$.
In this case, EM for SPNs generalizes EM for Gaussian mixture models.
If $\varphi_{l}\left(X_{l}\right)$ is a \emph{tree graphical model}
over discrete variables, the solution of eq. \ref{eq:aasda} can be
found with the Chow-Liu algorithm (\cite{ApproxDistrWithTrees}) adapted
for weighted likelihood (see \cite{LearningWithMixturesOfTrees}).
The algorithm has a cost quadratic on the cardinality of $X$ and
allows to learn jointly the optimal tree structure and potentials.

\section{Convergence for General Leaf Distributions }

The EM algorithm proceeds by iterating E-and-M steps (pseudocode in
Algorithm \ref{algEM}) until convergence. The training set log-likelihood
is guaranteed not to decrease at each step as long as the M-step maximization
can be done at least partially \cite{NealHinton98_incremEM}: namely,
calling $\theta_{l,new}$ and $\theta_{l,old}$ the current and previous
parameters of leaf $l$, this implies EM \emph{converges} if the update
at each leaf satisfies: 
\begin{equation}
\sum_{n=1}^{N}\alpha_{ln}\ln\varphi_{l}\left(x_{n}|\theta_{l,new}\right)\ge\sum_{n=1}^{N}\alpha_{ln}\ln\varphi_{l}\left(x_{n}|\theta_{l,old}\right)\label{eq:bound}
\end{equation}
This condition is very non-constraining, as it simply requires that
weighted log-likelihood can be at least approximately optimized. Note
that weighted log-likelihood maximization requires minor modifications
from standard maximum-likelihood. If approximate methods are used,
a simple check on bound (\ref{eq:bound}) ensures that the approximate
learning procedure did not decrease the lower bound (Algorithm \ref{algEM}
row $8$).

This allows a very \emph{broad family} of distributions to be used
as leaves: for instance, approximate maximum likelihood methods are
available for intractable graphical models (\cite{Wainwright:2008}),
probabilistic Neural Networks (\cite{Specht:1990}), probabilistic
Support Vector Machines (\cite{Platt99probabilisticoutputs}) and
several non parametric models (see e.g. \cite{Geman1982} and \cite{Cule2010}).
EM leaf distribution updates can be straightforwardly applied to each
of these models. Note that depending on the tractability of the leaf
distribution, some operations might not be tractable (e.q. exact marginalization
in general graphical models) - whether to use certain distributions
as leaves depends on the kind of queries one needs to answer and it
is an application specific decision. %It is hard to think of a probabilistic model that cannot be trained with maximum-likelihood at least approximately. 

%%If a procedure to maximize the log-likelihood exists then typically the weighted LL maximization requires small changes from non-weighted maximum likelihood cases (it corresponds basically to replicating some samples), therefore distributions that can be trained with maximum likelihood usually also the weighted variant. 

\textbf{Cost.} All the quantities required in a EM updates can be
computed with a single forward-downward pass on the SPN, thus an EM
iteration has cost \emph{linear in the number of edges}. The cost
of the maximization for each leaf depends on the leaf type, and it
is an application specific problem. For instance, with Gaussian and
tree graphical model leaves it is linear in the number of samples.

\begin{algorithm}
\caption{Compute $\alpha,\beta$ ($S,\{x_{1},x_{2},...,x_{N}\}$)\label{algAlphaU}}

\begin{algorithmic}[1]  
\label{algAlphaU} 
\State \textbf{Input:} SPN $S({W,\theta})$, samples $\{x_1,x_2,...,x_N\}$
\State set $\beta^q_i=0$ for each $(q,i) \in \mathcal{E} \left( S \right) $ %\State set $\alpha_{kn}=0$ for each leaf node $k$ and sample $n$ \For{each $x_{n} \in \{x_1,x_2,...,x_N\}$}  	\State compute $S_{k}(x_{n})$,$\frac{\partial S\left(x_{n}\right)}{\partial S_{k}}$ for each node $k$ in $S$  	\State (with a single forward-downward pass over $S(x_{n})$)
	\For{each sum node $q$ in $ S $ and each node $i \in ch(q)$}  		%\For{each node $i \in ch(q)$}  			\State $\beta^q_i \leftarrow \beta^q_i+\frac{1}{N}w_{i}^{q}S_{i}\left(x_{n}\right)\frac{\partial S\left(x_{n}\right)}{\partial S_{q}}S\left(x_{n}\right)^{-1}$ 		%\EndFor  	\EndFor 
	\For{each leaf node $l$ in $S$}  		\State $\alpha_{ln} \leftarrow S\left(x_{n}\right)^{-1}\frac{\partial S\left(x_{n}\right)}{\partial S_{l}}{S}_{l}\left(x_{n}\right)$ 	
\EndFor  
\EndFor 
\end{algorithmic}   
\end{algorithm}

\begin{algorithm}[tb]
\caption{EMstep($S,\{x_{1},x_{2},...,x_{N}\}$)\label{algEM}}

\begin{algorithmic}[1]
\State \textbf{Input:} SPN $S (W,\theta)$, samples $\{x_1,x_2,...,x_N\}$
%\State $[\alpha,\beta] \leftarrow$ \Call{ComputeAlphaAndBeta}{$ S,W,\theta,\{x_1,x_2,...,x_N\}$} %\State $W^* \leftarrow$ \CALL{WeightProjection}{$ S,U $} //update $W$
\State $[\alpha,\beta] \leftarrow$ Compute $\alpha,\beta \left( S, \{x_1,...,x_N  \} \right)$
\For{each sum node $q$ in $ S $ and each node $i \in ch(q)$}  \State $w^{q}_i \leftarrow \beta^q_i / \sum_{i \in ch\left(q\right)} \beta^q_i $ \EndFor 
\For{each leaf node $l$ in $S$}  \State $\theta_{l}\leftarrow \arg\max_{\theta_l}\sum_{n=1}^{N}\alpha_{ln}\ln\varphi_{l}\left(x_{n}|\theta_l\right)$  		\State if eq. \ref{eq:bound} is not satisfied, discard the update  \EndFor 
\end{algorithmic}  
\end{algorithm}

\section{Empirical Evaluation \label{sec:Application---Soft}}

The aim of this section is to evaluate the potential advantages of
learning the parameters of complex leaf distributions in a density
estimation setting.

\textbf{Setup.} As leaf distribution of choice we take tree graphical
models in which the M-step can be solved exactly (section \ref{subsec:Exponential-Family-Leaves}).
First we need to fix the SPN structure. In order to keep the focus
on parameter learning rather than structure learning we chose to use
the simplest structure learning algorithm (LearnSPN, \cite{SPNstructureLearning2013}),
and augment it to use tree leaves by simply adding a fixed number
of tree leaves to each generated sum node $q$. The tree leaves are
initialized as a mixture model over the data that was used to learn
the subnetwork rooted in $q$ (see \cite{SPNstructureLearning2013}
for details). To keep the models small and the structure simple, we
limit the depth to a fixed value. The number of added trees and the
maximum depth are hyperparameters.

\textbf{Methodology.} We evaluate the model on $20$ real-life datasets
for density estimation, whose structure is described in table \ref{tableLLres}
(see \cite{SPNstructureLearning2013}). These datasets are binary,
with a number of variables ranging from $16$ to $1556$, and have
been widely used as benchmark for density estimation (e.g. in \cite{ArithmCircuitsLearning},
\cite{SPNstructureLearning2013}, \cite{Rooshenas14}, \cite{Rahman16}).
We select the hyperparameters (described in \cite{SPNstructureLearning2013}
for details) performing a grid search over independence thresholds
(values $\{0.1,0.01,0.001\}$), number of tree leaves attached to
sum nodes ($\{5,20,30\}$), and maximum depth ${2,4,6}$. We train
$W,\theta$ with EM until validation log-likelihood convergence.

\textbf{Results.} We compare against two state-of-the-art parameter
learning methods: \emph{Concave-Convex Procedure} (CCCP, \cite{zhao16})
and \emph{Collapsed Variational Inference} (CVI, \cite{Zhao16collapsVarInf})
which employ LearnSPN for structure learning (like us, but without
depth limit) then re-learn the edge parameters of the resulting SPN.
The results of this experiment are shown in table \ref{tableLLres}.
To perform a fair comparison, we also plot the network size as the
number of edges in the network (table \ref{tableSpnSize}), and for
each tree leaf node we also add to this count the number of edges
which would be needed to represent the tree as a SPN. Our algorithm
(column TreeSPN) \emph{outperforms} CCCP and CVI in the majority of
cases, despite the network size being \emph{much smaller} (total number
of edges is $5.41$M vs. $27.1M$). These results indicates that it
can be convenient to use computational resources for modelling SPNs
with complex structured leaves, learned with EM, rather than in just
increasing the number of SPN edges. This new aspect should be explored
in future work.

%#############################################################################

\begin{table}[t]
\caption{Experimental results. Note that TreeSPN performs better ($11$ wins)
than both CCCP ($9$ wins) and CVI ($1$ win) while using much smaller
SPNs ($5.4M$ vs. $27M$ total edges).}

\label{tableSpnSize} \label{tableLLres}

\centering\resizebox{0.9\columnwidth}{!}{ %
\begin{tabular}{|l|ll|lll|ll|}
\hline 
 &  &  & \textbf{Test LL }  &  &  & \textbf{\#edges}  & \tabularnewline
Dataset  & Nvars  & $|$train$|$  & TreeSPN  & CCCP  & CVI  & TreeSPN  & CCCP \tabularnewline
\hline 
NLTCS  & $16$  & $16181$  & $\mathbf{-6.01}$  & $-6.03$  & $-6.08$  & $\mathbf{2K}$  & ${14K}$ \tabularnewline
MSNBC  & $17$  & $291326$  & $\mathbf{-6.04}$  & $-6.05$  & $-6.29$  & $\mathbf{13K}$  & ${55K}$ \tabularnewline
KDDCup2K  & $64$  & $180092$  & $-2.14$  & $\mathbf{-2.13}$  & $-2.14$  & ${50K}$  & $\mathbf{48K}$ \tabularnewline
Plants  & $69$  & $17412$  & $\mathbf{-12.30}$  & $-12.87$  & $-12.86$  & $\mathbf{60K}$  & ${133K}$ \tabularnewline
Audio  & $100$  & $15000$  & $\mathbf{-39.76}$  & $-40.02$  & $-40.6$  & $\mathbf{93K}$  & ${740K}$ \tabularnewline
Jester  & $100$  & $9000$  & $\mathbf{-52.59}$  & $-52.88$  & $-53.84$  & $\mathbf{93K}$  & ${314K}$ \tabularnewline
Netflix  & $100$  & $15000$  & $\mathbf{-56.12}$  & $-56.78$  & $-57.96$  & $\mathbf{94K}$  & ${162K}$ \tabularnewline
Accidents  & $111$  & $12758$  & $-29.86$  & $\mathbf{-27.70}$  & $-29.55$  & $\mathbf{100K}$  & ${205K}$ \tabularnewline
Retail  & $135$  & $22041$  & $-10.95$  & $-10.92$  & $\mathbf{-10.91}$  & ${116K}$  & $\mathbf{57K}$ \tabularnewline
Pumsb-star  & $163$  & $12262$  & $\mathbf{-23.71}$  & $-24.23$  & $-25.93$  & $\mathbf{105K}$  & ${140K}$ \tabularnewline
DNA  & $180$  & $1600$  & $\mathbf{-79.90}$  & $-84.92$  & $-86.73$  & ${167K}$  & $\mathbf{108K}$ \tabularnewline
Kosarek  & $190$  & $33375$  & $\mathbf{-10.75}$  & $-10.88$  & $-10.70$  & $\mathbf{149K}$  & ${203K}$ \tabularnewline
MSWeb  & $294$  & $29441$  & $-10.03$  & $\mathbf{-9.97}$  & $-9.89$  & ${186K}$  & $\mathbf{69K}$ \tabularnewline
Book  & $500$  & $8700$  & $\mathbf{-34.68}$  & $-35.01$  & $-34.44$  & ${434K}$  & $\mathbf{191K}$ \tabularnewline
EachMovie  & $500$  & $4524$  & $-55.42$  & $\mathbf{-52.56}$  & $-52.63$  & $\mathbf{339K}$  & ${523K}$ \tabularnewline
WebKB  & $839$  & $2803$  & $-167.8$  & $\mathbf{-157.5}$  & $-161.5$  & $\mathbf{713K}$  & ${1.44M}$ \tabularnewline
Reuters-52  & $889$  & $6532$  & $-91.69$  & $\mathbf{-84.63}$  & $-85.45$  & $\mathbf{604K}$  & ${2.21M}$ \tabularnewline
20Newsgrp.  & $910$  & $11293$  & $-156.8$  & $\mathbf{-153.2}$  & $-155.6$  & $\mathbf{848K}$  & ${14.6M}$ \tabularnewline
BBC  & $1058$  & $1670$  & $-266.3$  & $\mathbf{-248.6}$  & $-251.2$  & $\mathbf{881K}$  & ${1.88M}$ \tabularnewline
Ad  & $1556$  & $2461$  & $\mathbf{-16.88}$  & $-27.20$  & $-19.00$  & $\mathbf{364K}$  & ${4.13M}$ \tabularnewline
\hline 
\#Wins/TotSize  &  &  & $\mathbf{11}$  & $8$  & $1$  & $\mathbf{5.41}M$  & ${27.1M}$ \tabularnewline
\hline 
\end{tabular}} 
\end{table}

\section{Conclusions }

In this paper we derived the first parameter learning procedure for
SPNs which allows to train jointly edge weights and a wide class of
complex leaf distributions. Learning the leaf models corresponds to
fitting models with weighted maximum, and the algorithm converges
if this optimization is even partially performed. Experimental results
on $20$ datasets for density estimation showed that using complex
SPN leaves trained with EM produced better results than state-of-the-art
edge weights learning methods for SPNs while using much smaller models,
suggesting that learning complex SPN leaves is a promising direction
for future research.

\appendix
%dummy comment inserted by tex2lyx to ensure that this paragraph is not empty%dummy comment inserted by tex2lyx to ensure that this paragraph is not empty

\section{Proofs\label{appendix}}

\paragraph{Preliminars.}

Consider some subnetwork $\sigma_{c}$ of $S$ including the edge
$\left(q,i\right)$ (fig. \ref{fig:lemEx2}). Remembering that $\sigma_{c}$
is a tree, we divide $\sigma_{c}$ in three disjoint subgraphs: the
edge $\left(q,i\right)$, the tree $\sigma_{h\left(c\right)}^{\desc\left(i\right)}$
corresponding to ``descendants'' of $i$, and the remaining tree
$\sigma_{g\left(c\right)}^{\pred\left(q\right)}$. Notice that $g\left(c\right)$
could be the same for two different subnetworks $\sigma_{1}$and $\sigma_{2}$,
meaning that the subtree $\sigma_{g\left(c\right)}^{\pred\left(q\right)}$
is in common (similarly for $\sigma_{h\left(c\right)}^{\desc\left(i\right)}$).
We now observe that the the coefficient $\lambda_{c}$ and component
$P_{c}$ (def. \ref{def:pcLambda}) factorize in terms corresponding
to $\sigma_{g\left(c\right)}^{\pred\left(q\right)}$ and to $\sigma_{h\left(c\right)}^{\desc\left(i\right)}$
as follows: $\lambda_{c}=w_{i}^{q}\lambda_{h\left(c\right)}^{\desc\left(i\right)}\lambda_{g\left(c\right)}^{\pred\left(q\right)}$
and $P_{c}=P_{h\left(c\right)}^{\desc\left(i\right)}P_{g\left(c\right)}^{\pred\left(q\right)}$,
where $\lambda_{h\left(c\right)}^{\desc\left(i\right)}=\prod_{\left(m,n\right)\in\leaves\left(\sigma_{h\left(c\right)}^{\desc\left(i\right)}\right)}w_{n}^{m}$,
$P_{h\left(c\right)}^{\desc\left(i\right)}=\prod_{l\in\leaves\left(\sigma_{h\left(c\right)}^{\desc\left(i\right)}\right)}\varphi_{l}$
and similarly for $\pred\left(q\right)$. With this notation, for
\emph{each} subnetwork $\sigma_{c}$ including $\left(q,i\right)$
we write: 
\begin{equation}
\lambda_{c}P_{c}=w_{i}^{q}\left(\lambda_{g\left(c\right)}^{\pred\left(q\right)}P_{g\left(c\right)}^{\pred\left(q\right)}\right)\left(\lambda_{h\left(c\right)}^{\desc\left(i\right)}P_{h\left(c\right)}^{\desc\left(i\right)}\right)\label{eq:aboveBelow}
\end{equation}
Let us now consider the sum over \emph{all} the subnetworks $\sigma_{c}$
of $S$ that include $\left(q,i\right)$. The sum can be rewritten
as two nested sums, the external one over all terms $\sigma_{g}^{\pred\left(q\right)}$
(red part, fig. \ref{fig:lemEx2}) and the internal one over all subnets
$\sigma_{h}^{\desc\left(i\right)}$ (blue part, fig. \ref{fig:lemEx2}).
This is intuitively easy to grasp: we can think of the sum over all
trees $\sigma_{c}$ as first keeping the subtree $\sigma_{g}^{\pred\left(q\right)}$
fixed and varying all possible subtrees $\sigma_{h}^{\desc\left(i\right)}$
below $i$ (inner sum), then iterating this for choice of $\sigma_{g}^{\pred\left(q\right)}$
(outer sum). Exploiting the factorization \ref{eq:aboveBelow} we
obtain the following: 
\begin{equation}
\sum_{c:\left(q,i\right)\in\edg\left(\sigma_{c}\right)}\lambda_{c}P_{c}=w_{i}^{q}\sum_{g=1}^{C_{\pred\left(q\right)}}\lambda_{g}^{\pred\left(q\right)}P_{g}^{\pred\left(q\right)}\sum_{h=1}^{C_{\desc\left(i\right)}}\lambda_{h}^{\desc\left(i\right)}P_{h}^{\desc\left(i\right)}\label{eq:sumsigma}
\end{equation}
where $C_{\desc\left(i\right)}$ and $C_{\pred\left(q\right)}$ denote
the total number of different trees $\sigma_{h}^{\desc\left(i\right)}$
and $\sigma_{g}^{\pred\left(q\right)}$ in $S$. \begin{lem} \label{lem:Defining-as-}
$\frac{\partial S\left(X\right)}{\partial S_{q}}=\sum_{g=1}^{C_{\pred\left(q\right)}}\lambda_{g}^{\pred\left(q\right)}P_{g}^{\pred\left(q\right)}$.
\end{lem} \emph{Proof. }First let us separate the sum in eq. \ref{eq:spnmix}
in two sums, one over subnetworks including $q$ and one over subnetworks
not including $q$: $S\left(X\right)=\sum_{k:q\in\sigma_{k}}\lambda_{k}P_{k}+\sum_{l:q\notin\sigma_{l}}\lambda_{l}P_{l}$.
The second sum does not involve node $q$ so for $\frac{\partial S\left(X\right)}{\partial S_{q}}$
it is a constant $\hat{k}$. Then, $S=\sum_{k:q\in\sigma_{k}}\lambda_{k}P_{k}+\hat{k}$.
As in eq. \ref{eq:sumsigma}, we divide the sum $\sum_{k:q\in\sigma_{k}}\left(\cdot\right)$
in two nested sums acting over disjoint terms: 
\[
S=\left(\sum_{g=1}^{C_{\pred\left(q\right)}}\lambda_{g}^{\pred\left(q\right)}P_{g}^{\pred\left(q\right)}\right)\left(\sum_{h=1}^{C_{\desc\left(q\right)}}\lambda_{h}^{\desc\left(q\right)}P_{h}^{\desc\left(q\right)}\right)+\hat{k}
\]
. We now notice that $\sum_{k=1}^{C_{\desc\left(q\right)}}\lambda_{k}^{\desc\left(q\right)}P_{k}^{\desc\left(q\right)}=S_{q}$
by Proposition \ref{prop:A-SPN-S}, since $\lambda_{k}^{\desc\left(q\right)}P_{k}^{\desc\left(q\right)}$
refer to the subtree of $\sigma_{c}$ rooted in $i$ and the sum is
taken over all such subtrees. Therefore: $S=\left(\sum_{g=1}^{C_{\pred\left(q\right)}}\lambda_{g}^{\pred\left(q\right)}P_{g}^{\pred\left(q\right)}\right)S_{q}+\hat{k}$.
Taking the partial derivative leads to the result. \qed

\subsection{Proof of Proposition \ref{prop:Consider-a-SPN}\label{subsec:Proof-of-Lemma}}

We start by writing the sum on the left-hand side of eq. \ref{eq:sfsfr}
as in eq. \ref{eq:sumsigma}. Now, first we notice that $\sum_{k=1}^{C_{\desc\left(i\right)}}\lambda_{k}^{\desc\left(i\right)}P_{k}^{\desc\left(i\right)}$
equals $S_{i}\left(X\right)$ by Proposition \ref{prop:A-SPN-S},
since $\lambda_{k}^{\desc\left(i\right)}P_{k}^{\desc\left(i\right)}$
refer to the subtree of $\sigma_{c}$ rooted in $i$ and the sum is
taken over all such subtrees. Second, $\sum_{g=1}^{C_{\pred\left(q\right)}}\lambda_{g}^{\pred\left(q\right)}P_{g}^{\pred\left(q\right)}=\frac{\partial S\left(X\right)}{\partial S_{q}}$
for Lemma \ref{lem:Defining-as-}. Substituting in \ref{eq:sumsigma}
we get the result. \qed

\subsection{M-step for Edge Weights \label{subsec:EM-wDeriv}}

Starting from eq. \ref{eq:Q} and collecting terms not depending on
$W$ in a constant, we obtain: 
\begin{align*}
Q\left(W\right)= & \sum_{n=1}^{N}\sum_{c=1}^{C}\frac{\lambda_{c}P_{c}\left(x_{n}\right)}{S\left(x_{n}\right)}\ln\lambda_{c}\left(W\right)+\mbox{const}\\
= & \sum_{n=1}^{N}\sum_{c=1}^{C}\frac{\lambda_{c}P_{c}\left(x_{n}\right)}{S\left(x_{n}\right)}\sum_{\left(q,i\right)\in\edg\left(\sigma_{c}\right)}\ln w_{i}^{q}+\mbox{const}
\end{align*}
We now drop the constant and move out $\sum_{\left(q,i\right)\in\edg\left(\sigma_{c}\right)}$
by introducing $\delta_{\left(q,i\right),c}$ s.t. $\delta_{\left(q,i\right),c}=1$
if $\left(q,i\right)\in\edg\left(\sigma_{c}\right)$ and $0$ otherwise
and summing over \emph{all} edges $\edg\left(S\right)$: 
\begin{align*}
Q\left(W\right)= & \sum_{n=1}^{N}\sum_{c=1}^{C}\frac{\lambda_{c}P_{c}\left(x_{n}\right)}{S\left(x_{n}\right)}\sum_{\left(q,i\right)\in\edg\left(S\right)}\ln w_{i}^{q}\delta_{\left(q,i\right),c}\\
= & \sum_{\left(q,i\right)\in\edg\left(S\right)}\sum_{n=1}^{N}\frac{\sum_{c=1}^{C}\lambda_{c}P_{c}\left(x_{n}\right)\delta_{\left(q,i\right),c}}{S\left(x_{n}\right)}\ln w_{i}^{q}\\
= & \sum_{\left(q,i\right)\in\edg\left(S\right)}\sum_{n=1}^{N}\frac{\sum_{c:\left(q,i\right)\in\edg\left(\sigma_{c}\right)}\lambda_{c}P_{c}\left(x_{n}\right)}{S\left(x_{n}\right)}\ln w_{i}^{q}
\end{align*}
Applying Proposition \ref{prop:Consider-a-SPN} we get: $Q\left(W\right)=\sum_{\left(q,i\right)\in\edg\left(S\right)}\left(\sum_{n=1}^{N}\frac{w_{i,old}^{q}\frac{\partial S\left(x_{n}\right)}{\partial S_{q}}S_{i}\left(x_{n}\right)}{S\left(x_{n}\right)}\right)\ln w_{i}^{q}$,
and defining $\beta_{i}^{q}=w_{i,old}^{q}\sum_{n=1}^{N}S\left(x_{n}\right)^{-1}\frac{\partial S\left(x_{n}\right)}{\partial S_{q}}S_{i}\left(x_{n}\right)$
we write $Q\left(W\right)=\sum_{q\in\sums\left(S\right)}\sum_{i\in ch(q)}\beta_{i}^{q}\ln w_{i}^{q}$.
\qed

\subsection{M-step for Leaf Distributions \label{subsec:EM-thetaDeriv}}

Starting from eq. \ref{eq:Q}, as in \ref{subsec:EM-wDeriv} we expand
$\ln P_{c}$ as a sum of logarithms and obtain: 
\[
Q\left(\theta\right)=\sum_{n=1}^{N}\sum_{c=1}^{C}\frac{\lambda_{c}P_{c}\left(x_{n}\right)}{S\left(x_{n}\right)}\sum_{l\in\leaves\left(\sigma_{c}\right)}\ln\varphi_{l}\left(x_{n}|\theta_{l}\right)+const.
\]
Introducing $\delta_{l,c}$ which equals $1$ if $l\in\leaves\left(\sigma_{c}\right)$
and $0$ otherwise, dropping the constant and performing the sum $\sum_{l\in\leaves\left(S\right)}$
over all leaves in $S$ we get: 
\begin{align*}
Q\left(\theta\right)= & \sum_{n=1}^{N}\sum_{c=1}^{C}\frac{\lambda_{c}P_{c}\left(x_{n}\right)}{S\left(x_{n}\right)}\sum_{l\in\leaves\left(S\right)}\ln\varphi_{l}\left(x_{n}|\theta_{l}\right)\delta_{l,c}\\
= & \sum_{l\in\leaves\left(S\right)}\sum_{n=1}^{N}\frac{\sum_{c=1}^{C}\lambda_{c}P_{c}\left(x_{n}\right)}{S\left(x_{n}\right)}\ln\varphi_{l}\left(x_{n}|\theta_{l}\right)\delta_{l,c}\\
= & \sum_{l\in\leaves\left(S\right)}\sum_{n=1}^{N}\frac{\sum_{c:l\in\leaves\left(\sigma_{c}\right)}\lambda_{c}P_{c}\left(x_{n}\right)}{S\left(x_{n}\right)}\ln\varphi_{l}\left(x_{n}|\theta_{l}\right)\\
= & \sum_{l\in\leaves\left(S\right)}\sum_{n=1}^{N}\alpha_{ln}\ln\varphi_{l}\left(x_{n}|\theta_{l}\right)
\end{align*}
Where $\alpha_{ln}=S\left(x_{n}\right)^{-1}\sum_{c:l\in\leaves\left(\sigma_{c}\right)}\lambda_{c}P_{c}\left(x_{n}\right)$.
To compute $\alpha_{ln}$ we notice that the term $P_{c}$ in this
sum always contains a factor $\varphi_{l}$ (def. \ref{def:pcLambda}),
and $\varphi_{l}=S_{l}$ by def. \ref{def:SPN}. Then, writing $P_{c\backslash l}=\left(\prod_{k\in\leaves\left(\sigma_{c}\right)\backslash l}\varphi_{k}\right)$
we obtain: $\alpha_{ln}=S\left(x_{n}\right)^{-1}S_{l}\left(\sum_{c:l\in\leaves\left(\sigma_{c}\right)}\lambda_{c}P_{c\backslash l}\left(x_{n}\right)\right)$.
Finally, since $S=\sum_{c:l\in\leaves\left(\sigma_{c}\right)}\lambda_{c}P_{c}+\sum_{k:l\notin\leaves\left(\sigma_{k}\right)}\lambda_{k}P_{k}=S_{l}\sum_{c:l\in\leaves\left(\sigma_{c}\right)}\lambda_{c}P_{c\backslash l}+\hat{k}$
(where $\hat{k}$ does not depend on $S_{l}$), taking the derivative
we get $\frac{\partial S}{\partial S_{l}}=\sum_{c:l\in\leaves\left(\sigma_{c}\right)}\lambda_{c}P_{c\backslash l}$.
Substituting we get: $\alpha_{ln}=S\left(x_{n}\right)^{-1}\frac{\partial S\left(X\right)}{\partial S_{l}}S_{l}\left(x_{n}\right)$.
\qed

 \bibliographystyle{apalike}
\bibliography{bibliography}

\end{document}